\documentclass[conference]{IEEEtran}
\IEEEoverridecommandlockouts
\usepackage{cite}
\usepackage{amsmath,amssymb,amsfonts}
\usepackage{algorithmic}
\usepackage{graphicx}
\usepackage{textcomp}
\usepackage{xcolor}
\usepackage{spconf,amsmath,graphicx}
\usepackage{multirow}
\usepackage{tabularray}
\usepackage{subfig}
\usepackage{algorithm}
\usepackage{algorithmic}
\usepackage{amssymb}
\usepackage{booktabs}
\usepackage{colortbl}
\usepackage{pifont}       % \ding{xx}
\usepackage{bbding}       % \Checkmark,\XSolid,... (需要和pifont宏包共同使用)
\usepackage{fontawesome}  % \faCheck,\faTimes
\usepackage{tabularx}
\usepackage{url}
\def\BibTeX{{\rm B\kern-.05em{\sc i\kern-.025em b}\kern-.08em
    T\kern-.1667em\lower.7ex\hbox{E}\kern-.125emX}}

\usepackage{svg}

\begin{document}

\title{Enhanced Cross-modal 3D Retrieval via \\
Tri-modal Reconstruction
\thanks{
%\vspace{-1.0em}
* Corresponding author. 

This research is supported by the National Natural Science Foundation of China (No. 62406267), Guangzhou-HKUST(GZ) Joint Funding Program (Grant No.2025A03J3956), Education Bureau of Guangzhou Municipality and the Guangzhou Municipal Education Project (No. 2024312122).
}
}

%\author{Anonymous ICME submission}
\author{
\IEEEauthorblockN{Junlong Ren, Hao Wang\textsuperscript{*}}
\IEEEauthorblockA{\textit{The Hong Kong University of Science and Technology (Guangzhou)}, Guangzhou, China \\
Email: jren686@connect.hkust-gz.edu.cn, haowang@hkust-gz.edu.cn}
}

\maketitle

\begin{abstract}

Cross-modal 3D retrieval is a critical yet challenging task, aiming to achieve bi-directional retrieval between 3D and text modalities. 
Current methods predominantly rely on a certain 3D representation (e.g., point cloud), with few exploiting the 2D-3D consistency and complementary relationships, which constrains their performance. To bridge this gap, we propose to adopt multi-view images and point clouds to jointly represent 3D shapes, facilitating tri-modal alignment (i.e., image, point, text) for enhanced cross-modal 3D retrieval.
Notably, we introduce tri-modal reconstruction to improve the generalization ability of encoders. Given point features, we reconstruct image features under the guidance of text features, and vice versa.
With well-aligned point cloud and multi-view image features, we aggregate them as multimodal embeddings through fine-grained 2D-3D fusion to enhance geometric and semantic understanding.
Recognizing the significant noise in current datasets where many 3D shapes and texts share similar semantics, we employ hard negative contrastive training to emphasize harder negatives with greater significance, leading to robust discriminative embeddings.
Extensive experiments on the Text2Shape dataset demonstrate that our method significantly outperforms previous state-of-the-art methods in both shape-to-text and text-to-shape retrieval tasks by a substantial margin.

\end{abstract}


\section{Introduction}
\label{sec:intro}

The perception and understanding of the 3D world play a crucial role in robotics, spatial intelligence, etc. Since natural language provides an intuitive method for interacting with the 3D world, it is essential to bridge the gap between 3D and textual modalities. In particular, cross-modal 3D retrieval, which aims to achieve bi-directional retrieval between 3D shapes and texts, emerges as a crucial task.

Previous works on cross-modal 3D retrieval \cite{chen2019text2shape, han2019y2seq2seq, ruan2024tricolo, tang2021parts2words, wu2024com3d} primarily focus on learning a joint embedding space for vision-language modalities. 
However, most of these methods merely employ a certain modality to represent 3D shapes, such as point clouds \cite{tang2021parts2words}. 
Relying on only one modality limits the model's ability to capture the full range of geometric and semantic information.
Although \cite{ruan2024tricolo,wu2024com3d} adopt multiple 3D representations (e.g., 2D images and 3D voxels), they do not exploit the complementary relationships between these modalities, failing to fully leverage the 2D-3D consistency.

\begin{figure}[t]
\vspace{-0.85em}
\centering
%\captionsetup{justification=centering} % 设置居中对齐
\setlength{\abovecaptionskip}{0pt} % 减少标题上方的间距
\setlength{\intextsep}{0pt} % 减少文本与图片之间的间距
\setlength{\textfloatsep}{0pt} % 减少文本与浮动对象之间的间距

\resizebox{0.9\linewidth}{!}{
  \subfloat{\hspace{-2.8mm} \includegraphics[width=0.50\linewidth]{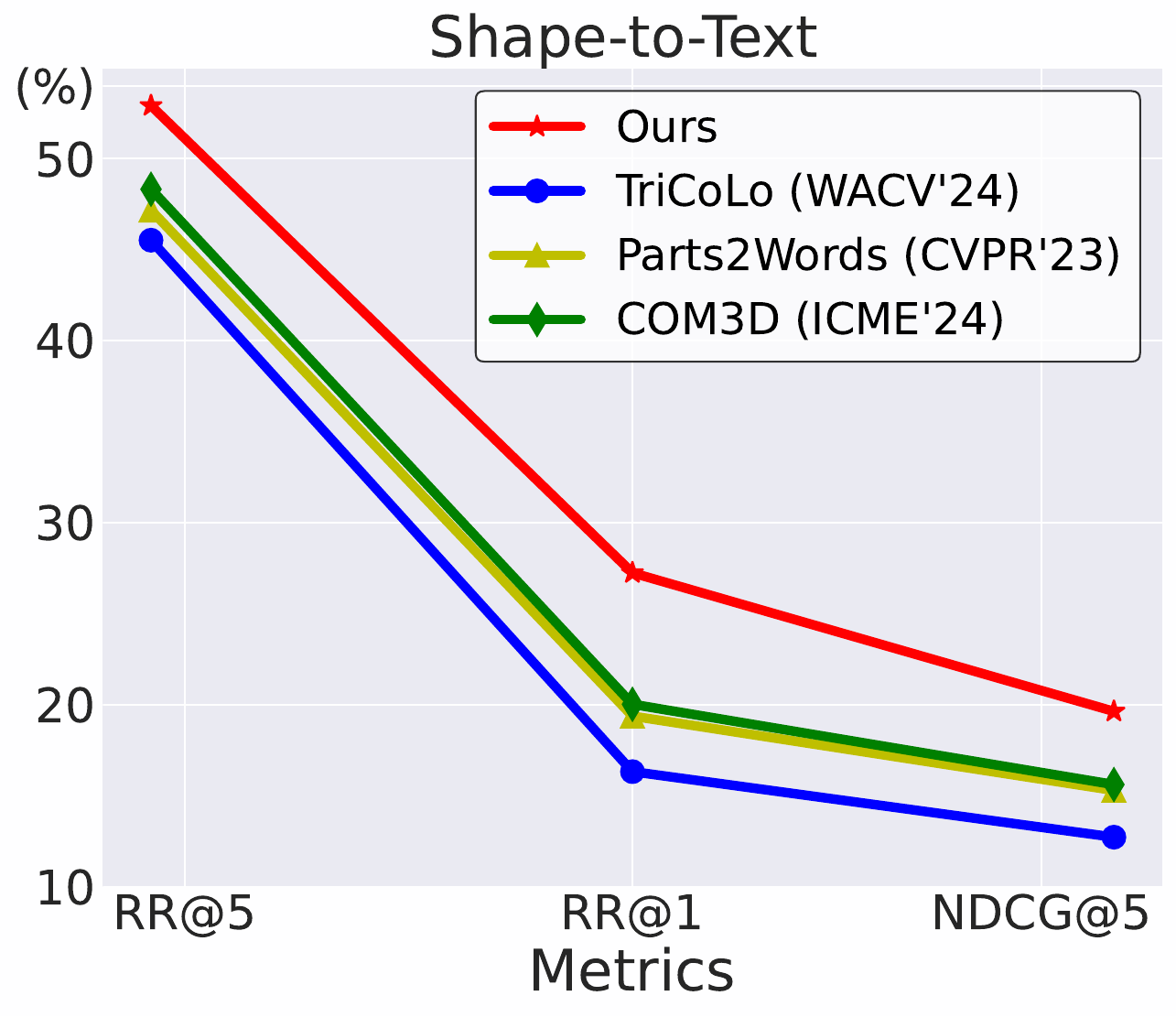}}
  \hfill
  \subfloat{\includegraphics[width=0.50\linewidth]{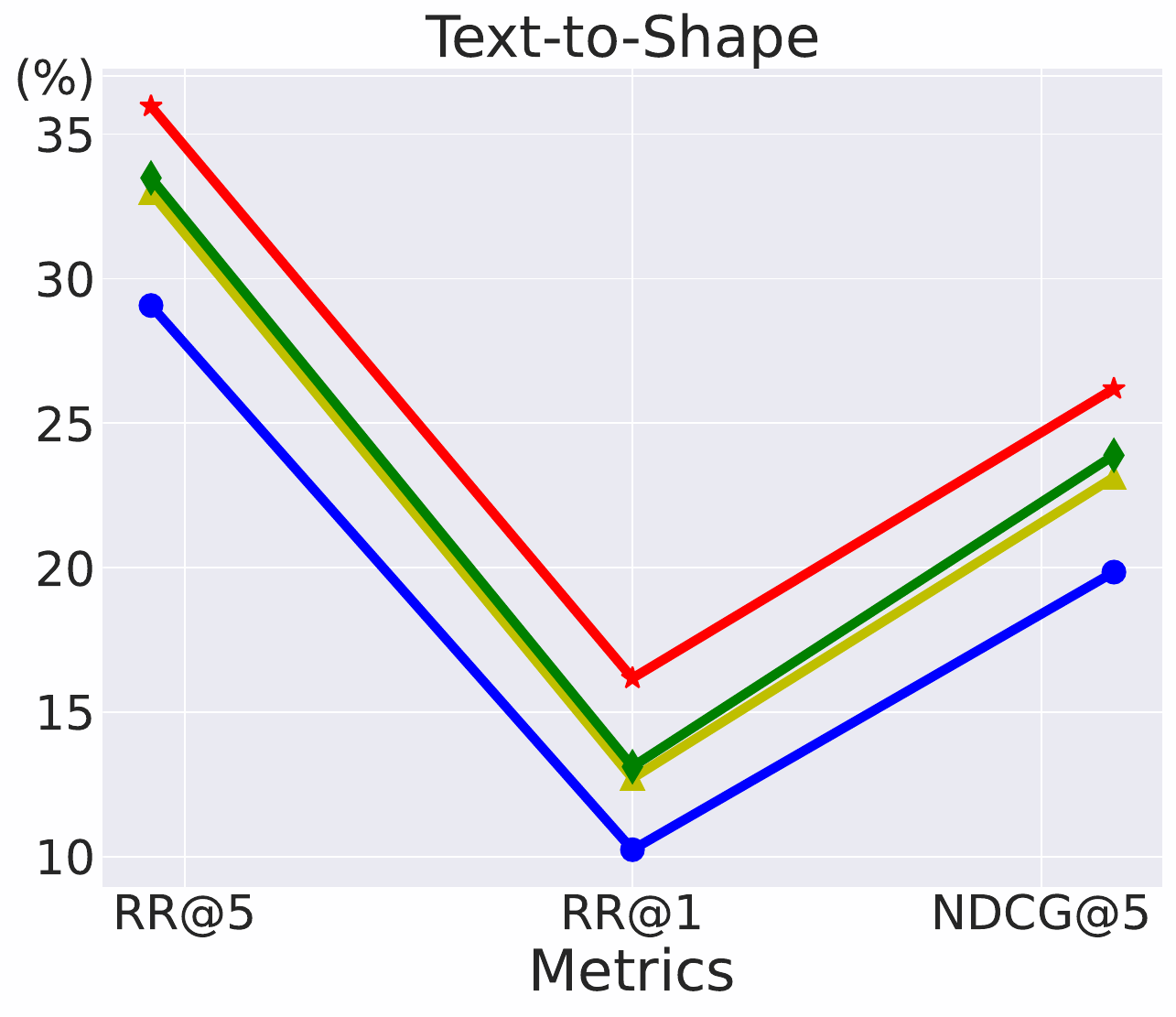}}
}
\caption{\textbf{Comparison with previous methods on shape-to-text and text-to-shape retrieval.} We outperform these works by a large margin over all metrics on the Text2Shape dataset \cite{chen2019text2shape}.}
\label{fig:s2t_and_t2s_performance}
\vspace{-1.5em}
\end{figure}

Specifically, TriCoLo \cite{ruan2024tricolo} aligns the image-text and voxel-text pairs separately, in which the complementary relationships between image-voxel modalities are not fully explored. COM3D \cite{wu2024com3d} adopts Parts2Words \cite{tang2021parts2words} as its baseline and further fuses point and image features, but it does not align image-point modalities, limiting the fusion effect. 
As shown in Fig. \ref{fig:s2t_and_t2s_performance}, TriCoLo achieves the lowest accuracy because of its limited alignment on image-voxel modalities; similarly, COM3D achieves little improvement compared to Parts2Words.

To bridge the gap between 2D-3D-text tri-modal data, we propose to leverage the 2D-3D consistency and facilitate alignment among point, image, and text modalities. To be specific, this paper proposes a tri-modal reconstruction framework, where we aim to pull the multimodal embeddings from the same 3D shapes closer and enhance the generalization of encoders. Technically, we reconstruct image features with point features under the guidance of text features, and vice versa.
By involving all of the three modalities simultaneously in the reconstruction process, our model learns a comprehensive representation that captures the interrelationships between 2D images, 3D shapes, and textual descriptions.

Then, we exploit the complementary relationships of multi-view images and point clouds, which respectively contain dense semantic information and represent key 3D features such as spatial hierarchy and geometry.
To further promote the tri-modal alignment of the reconstruction process, we propose a fine-grained 2D-3D fusion module that aggregates well-aligned image and point features as multimodal embeddings.

Lastly, it is observed that numerous 3D shapes and texts exhibit similar semantic characteristics, which may confuse the tri-modal reconstruction and introduce noise.
To address this issue, we adopt hard negative contrastive training, emphasizing harder negatives with higher importance. This method leads to robust and discriminative embedding learning, by improving the cross-modal alignment in tri-modal reconstruction.

\begin{figure*}[!t]
\centering
  \subfloat{\includegraphics[width=1\linewidth]{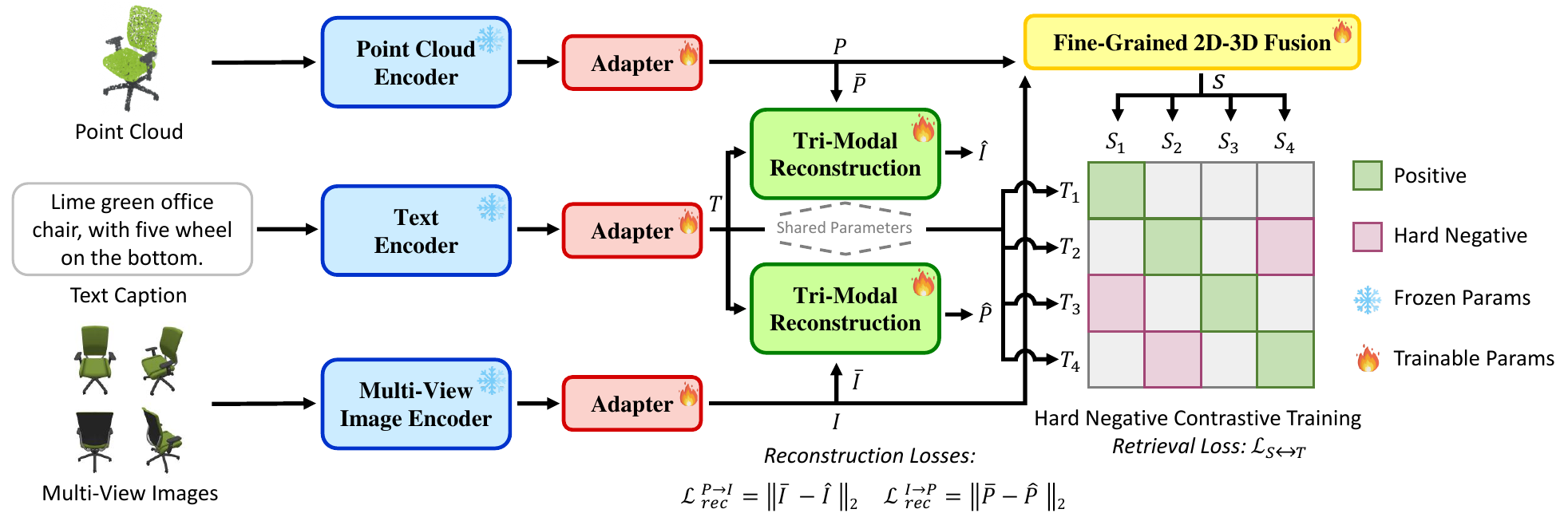}} 
   
\caption{\textbf{The overview of our proposed method.} It consists of three components: frozen encoders with trainable adapters for three modalities, tri-modal reconstruction, and fine-grained 2D-3D fusion. Each 3D shape is represented as a point cloud and multi-view images to utilize 2D-3D consistency and complementary relationships. \textbf{Tri-modal reconstruction} aims to reconstruct image features with point features under the guidance of text features, and vice versa. \textbf{Fine-grained 2D-3D fusion} aggregates point and image features to holistically represent 3D shapes. \textbf{Hard negative contrastive training} re-weights harder negatives with higher importance to learn and align discriminative embeddings.}
\label{figure::overview}
\vspace{-1.5em}
\end{figure*}

Extensive experiments and ablation studies verify the effectiveness of our method. As demonstrated in Fig. \ref{fig:s2t_and_t2s_performance}, our method surpasses existing methods on the Text2Shape dataset \cite{chen2019text2shape} by a substantial margin, achieving state-of-the-art results. In summary, our principal contributions are delineated as follows:
\begin{itemize}
    \item We propose a novel cross-modal 3D retrieval model with tri-modal reconstruction, prompting alignment among points, images, and texts to enhance generalization.
    \item  We introduce fine-grained 2D-3D fusion to bridge the gap between 2D and 3D visual information,  exploiting the 2D-3D consistency and complementary relationships.
    \item We employ hard negative contrastive training to improve the learning efficacy of discriminative embeddings by eliminating the noise in datasets.
\end{itemize}

\section{Related Work}

\subsection{2D-Text Retrieval}

The emergence of large-scale 2D-text pre-training datasets \cite{schuhmann2022laion5b,kakaobrain2022coyo-700m,miech2019howto100m} has significantly advanced the development of 2D-text retrieval. CLIP \cite{radford2021learning} stands as a milestone in recent years, being pre-trained on 400M image-text pairs from the web and achieving remarkable zero-shot performance across numerous datasets. It also serves as the foundation model for extensive downstream tasks, including multimodal large language models \cite{li2023blip2,huang2023language,liu2024visual,li-etal-2024-groundinggpt}. 
Follow-up works \cite{fang2023eva,li2022blip,zhai2023sigmoid,chen2023altclip} adopt optimized pre-training objectives for better performance and robustness.

\subsection{3D-Text Retrieval}

Text2Shape \cite{chen2019text2shape} represents a pioneering effort in cross-modal 3D retrieval by introducing a cross-modal 3D-text dataset. The collected Text2Shape dataset is the subset of ShapeNet \cite{chang2015shapenet} with additional textual descriptions. It also proposes a straightforward framework using 3D-CNN \cite{tran2015learning} and GRU \cite{chung2014empirical} to align 3D voxels with texts.
$\rm Y^{2}$Seq2Seq \cite{han2019y2seq2seq} mitigates the computational costs associated with the cubic complexity of 3D voxels by representing 3D shapes as multi-view images. TriCoLo \cite{ruan2024tricolo} introduces a contrastive training framework without complex attention mechanisms or losses. Parts2Words \cite{tang2021parts2words} segments point clouds into parts and employs regional-based matching between parts from shapes and words from texts. COM3D \cite{wu2024com3d} generates cross-view correspondence 3D features using a scene representation transformer \cite{sajjadi2022scene}.

\section{Method}

%\subsection{Framework Overview}
The overview of our proposed framework is illustrated in Fig. \ref{figure::overview}. The framework consists of three components: encoders for three modalities, a tri-modal reconstruction module, and a fine-grained 2D-3D fusion module. For a 3D shape associated with multiple modality representations (point cloud, multi-view images, and text), we extract embeddings for each modality through the corresponding encoder. 
Next, the tri-modal reconstruction module is employed to pull the multimodal embeddings from the same
3D shapes closer and promote the generalization ability of encoders.
Then, the fine-grained 2D-3D fusion module is adopted to obtain a unified 3D shape embedding with rich semantic and geometry information. 
Finally, we utilize hard negative contrastive training to learn discriminative 3D shapes and text embeddings for alignment.

\subsection{Multimodal Embeddings}

\paragraph{Point Cloud Encoder}
The point cloud encoder utilizes a frozen pre-trained Point-BERT \cite{yu2022point} as the backbone. 
It takes a point cloud $p \in \mathbb{R}^{n_p \times d_p}$ as input, where $n_p$ is the number of points and $d_p$ is the dimension of each point. The extracted point features are subsequently processed by the adapter to map point features to the high-level semantic space of vision-language modalities.
The encoded embeddings are denoted as $P = \left\{ p_{n} \right\}_{n = 1}^{N} \in \mathbb{R}^{N \times D}$, where $N$ is the number of point features and $D$ is the feature dimension.

\paragraph{Multi-View Image Encoder}
Each 3D shape is rendered into multi-view images using meshes to utilize the dense semantic information in the image modality. These images are encoded through a frozen CLIP \cite{radford2021learning} image encoder and a trainable adapter. The image features are represented as $I = \left\{ i_{m} \right\}_{m = 1}^{M} \in \mathbb{R}^{M \times D}$, where $M$ is the number of views.

\paragraph{Text Encoder}
The textual captions of 3D shapes are first processed through a frozen CLIP text encoder, followed by a trainable adapter. The extracted sentence-level features are denoted as $T \in \mathbb{R}^{D}$.

\subsection{Tri-Modal Reconstruction}
\begin{figure}[t]
\centering
% %\captionsetup{justification=centering} % 设置居中对齐
% \setlength{\abovecaptionskip}{0pt} % 减少标题上方的间距
% % \setlength{\belowcaptionskip}{0pt} % 减少标题下方的间距
% \setlength{\intextsep}{0pt} % 减少文本与图片之间的间距
% \setlength{\textfloatsep}{0pt} % 减少文本与浮动对象之间的间距
  \subfloat{\includegraphics[width=0.75\linewidth]{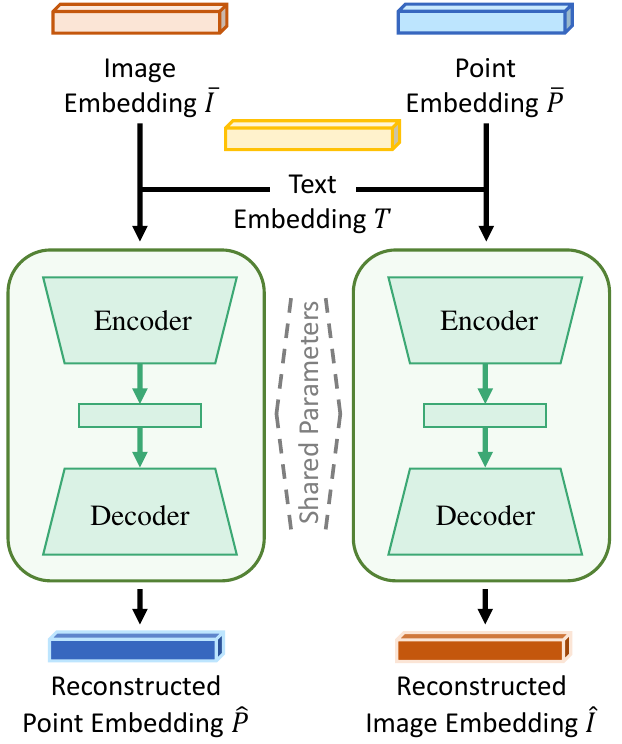}} 
   
\caption{\textbf{The pipeline of tri-modal reconstruction.} We reconstruct point embeddings using image and text embeddings and simultaneously reconstruct image embeddings with point and text embeddings.}
\label{figure::text guided reconstruction}
\vspace{-1.5em}
\end{figure}

To pull the multimodal embeddings from the same 3D shapes closer and enhance the generalization ability of encoders, we introduce tri-modal reconstruction.
Notably, we do not adopt the conventional bi-reconstruction \cite{feng2023hypergraph} which reconstructs features using only a single modality as input. Instead, we facilitate alignment among point, image, and text modalities simultaneously in the reconstruction module. 
By leveraging the complementary relationships and cross-modal consistency of point, image, and text features, our method improves the alignment and generalization across modalities.
Our proposed tri-modal reconstruction pipeline is displayed in Fig. \ref{figure::text guided reconstruction}.
Given point features as input, we reconstruct image features under the guidance of text features, and vice versa.
Note that we do not reconstruct text features since each 3D shape is associated with diverse text semantics in the datasets. Direct reconstruction of text features could cause model confusion and introduce noise into the training process.
We first pool the point and image features as $\bar{P}\in \mathbb{R}^{D}$ and $ \bar{I} \in \mathbb{R}^{D}$.
Then we concatenate $\bar{P}$ or $\bar{I}$ with text embeddings $T$ and reconstruct target modalities:
\begin{equation}
\resizebox{0.91\linewidth}{!}{
$
    \hat{I} = {MLP\left( \lbrack \bar{P};T\rbrack \right)} \in \mathbb{R}^{D},~\hat{P} = {MLP\left( \lbrack \bar{I};T\rbrack \right)} \in \mathbb{R}^{D},
    $
}
\end{equation}
where $MLP$ is a multi-layered perceptron (MLP) with ReLU activation function, $\hat{I}$ and $\hat{P}$ are the reconstructed image and point features, respectively. The image-to-point and point-to-image reconstruction restrictions are formulated as follows:

\begin{equation}
%\resizebox{1\linewidth}{!}{
%$
    \mathcal L_{rec}^{I \rightarrow P} = \left\|\bar{P}-\hat{P}\right\|_{2},~\mathcal L_{rec}^{P \rightarrow I} = \left\|\bar{I}-\hat{I}\right\|_{2},
%$
%}
\end{equation}
where $\left\| \cdot \right\|_{2}$ is the $\mathcal L_{2}$ norm.

%reconstruction restriction

\subsection{Fine-Grained 2D-3D Fusion}
To leverage collaborative information across 2D and 3D modalities, we fuse the image and point features as holistic 3D features. 
Concretely, we obtain the multimodal 3D features by applying the context-query attention \cite{yu2018qanet}, which models the fine-grained cross-modal interactions between point and image features for semantic fusion and alignment. We first calculate the similarity matrix $\mathcal{S} \in \mathbb{R}^{N \times M}$ between each point feature and image feature through the trilinear function \cite{seo2016bidirectional}:
\begin{equation}
    f\left( {p_{n},i_{m}} \right) = W_{0}\left\lbrack p_{n};i_{m};p_{n}\odot i_{m} \right\rbrack,
\end{equation}
where $W_{0} \in \mathbb{R}^{3D}$ is a learnable weight and $\odot$ is the element-wise multiplication. Then we compute two attention weights:
\begin{equation}
    \mathcal{A} = \mathcal{S}_{r} \cdot I \in \mathbb{R}^{N \times D},~\mathcal{B} = \mathcal{S}_{r} \cdot \mathcal{S}_{c}^{T} \cdot P \in \mathbb{R}^{N \times D},
\end{equation}
where $\mathcal{S}_r$ and $\mathcal{S}_c$ are the row-wise and column-wise normalized matrix of $S$ by Softmax, respectively. Finally, the encoded 3D shape embeddings are represented as:
\begin{equation}
    S = {PoolMLP\left( \lbrack P;\mathcal{A};P\odot \mathcal{A};P\odot \mathcal{B}\rbrack \right)} \in \mathbb{R}^{D},
\end{equation}
where $PoolMLP$ is an MLP with max-pooling.

\subsection{Training Objectives}

\paragraph{Hard Negative Contrastive Training}
Current datasets contain significant noise since numerous 3D shapes and texts share similar semantics, confusing the tri-modal reconstruction.
We then adopt the hard negative noise-contrastive estimation (HN-NCE) \cite{Radenovic_2023_CVPR} to eliminate the noise and learn discriminative 3D and text embeddings.
Different from vanilla InfoNCE \cite{oord2018representation} that uniformly sample negative samples, HN-NCE emphasizes harder negatives with higher importance, leading to better robustness. Concretely, negative samples within a batch are first re-weighted as:

\begin{table*}[t]     
\centering
  \caption{\textbf{Comparison results on the Text2shape dataset.} S2T and T2S indicate shape-to-text and text-to-shape retrieval, respectively. We achieve state-of-the-art results across all metrics.}
  \label{table::compare_with_SOTA}
\begin{tabular}{lccccccc}
\toprule
\multicolumn{1}{l}{Method} & \multicolumn{1}{c}{Venue} & \multicolumn{3}{c}{S2T} & \multicolumn{3}{c}{T2S} \\ 
\cmidrule(lr){3-5} \cmidrule(lr){6-8} 
                     &      &   RR@1   &  RR@5  & NDCG@5       & RR@1   &  RR@5  & NDCG@5      \\ \midrule
 Text2Shape \cite{chen2019text2shape}     &  ACCV'2018 & 0.83        &  3.37   & 0.73   &  0.40   &  2.37 & 1.35        \\
 $\rm Y^{2}$Seq2Seq \cite{han2019y2seq2seq}     & AAAI'2019 &  6.77         &  19.30     & 5.30 & 2.93    & 9.23    & 6.05    \\
 TriCoLo \cite{ruan2024tricolo}     & WACV'2024 & 16.33          & 45.52     & 12.73  & 10.25   & 29.07     &  19.85    \\
 Parts2Words \cite{tang2021parts2words}  &  CVPR'2023 & 19.38     &   47.17          & 15.30 & 12.72   & 32.98        & 23.13  \\
 COM3D \cite{wu2024com3d} &  ICME'2024  & 20.03     &   48.32          & 15.62 & 13.12   & 33.48    &   23.89    \\
 %\midrule
 \rowcolor{orange!15}  Ours  &  ICME'2025  &  \textbf{27.25}    &   \textbf{52.91}   &    \textbf{19.64}      & \textbf{16.18}   & \textbf{35.96}   &   \textbf{26.19}      \\
\bottomrule
\end{tabular}
\vspace{-1.5em}
\end{table*}

\begin{equation}
w_{i, j}=\frac{(n-1) \cdot e^{\beta Sim\left(S_i, T_j\right) / \tau}}{\sum_{k \neq i} e^{\beta Sim\left(S_i, T_k\right) / \tau}},
\end{equation}
where $n$ is the batch size, $\beta$ is the concentration parameter and $\tau$ is a learnable temperature parameter. $Sim\left(\cdot\right)$ denotes the similarity function, we utilize the cosine similarity:
\begin{equation}
    Sim\left(S_i, T_j\right)=\frac{S_i^{\top} T_j}{\left\|S_i\right\| \cdot\left\|T_j\right\|},
\end{equation}
where $S_i$ and $T_j$ are the $i$-th and $j$-th 3D shape and text embeddings within a batch, respectively.
The bi-directional shape-to-text and text-to-shape retrieval losses are as:

\begin{equation}
\resizebox{1\linewidth}{!}{
$
\begin{aligned}
\mathcal L_{S\leftrightarrow T}= & -\sum_{i=1}^{n} \log \left(\frac{e^{Sim\left(S_i, T_i\right) / \tau}}{e^{Sim\left(S_i, T_i\right) / \tau}+\sum_{j \neq i} e^{Sim\left(S_i, T_j\right) / \tau} w_{i, j}}\right) \\
& -\sum_{i=1}^{n} \log \left(\frac{e^{Sim\left(S_i, T_i\right) / \tau}}{e^{Sim\left(S_i, T_i\right) / \tau}+\sum_{j \neq i} e^{Sim\left(S_j, T_i\right) / \tau} w_{j, i}}\right).
\end{aligned}
$
}
\end{equation}

\paragraph{Overall Loss Formulation}
The final training objective is the sum of the retrieval and reconstruction losses to closely align 3D shapes and text captions:
\begin{equation}
\mathcal L = \mathcal L_{S\leftrightarrow T} + \mathcal L_{rec}^{I \rightarrow P} + \mathcal L_{rec}^{P \rightarrow I}.
\end{equation}

\section{Experiments}

\subsection{Experimental Setup}

\paragraph{Dataset}

We utilize the Text2Shape \cite{chen2019text2shape} dataset, commonly used in prior works. Text2Shape is a cross-modal dataset that includes 3D shapes with corresponding text captions. On average, each 3D shape has five textual descriptions, allowing the model to align 3D shapes with diverse text semantics. Following the data split defined by \cite{tang2021parts2words}, the training set consists of 11,498 3D shapes (57,538 3D-text pairs), while the test set contains 1,434 3D shapes (7,128 3D-text pairs).

\begin{table}[t]     
\scriptsize
\centering
\caption{\textbf{Ablation study on backbones.} In the first row, we adopt the same backbones as \cite{ruan2024tricolo, tang2021parts2words, wu2024com3d} for a fair comparison.}
\label{table::ablation study backbone}
\setlength{\tabcolsep}{3.5pt}
\resizebox{1\linewidth}{!}{
\begin{tabular}{ccccccccc}
\toprule
\multicolumn{3}{c}{Settings}   & \multicolumn{3}{c}{S2T} & \multicolumn{3}{c}{T2S} \\ 
 \cmidrule(lr){1-3} \cmidrule(lr){4-6} \cmidrule(lr){7-9} 
  Image & Point & Text    &   RR@1   &  RR@5  & NDCG@5       & RR@1   &  RR@5  & NDCG@5      \\ \midrule
MVCNN & PointNet &  GRU &      25.40 & \textbf{55.07} & 18.73 & 14.03 & 34.85 & 24.76      \\
\rowcolor{orange!15}  CLIP & Point-BERT    & CLIP &    \textbf{27.25}    &   52.91   &    \textbf{19.64}      & \textbf{16.18}   & \textbf{35.96}   &   \textbf{26.19}      \\

\bottomrule
\end{tabular}
}
\vspace{-0.5em}
\end{table}

\begin{table}[t]     
\scriptsize

\centering
\caption{\textbf{Ablation study on input modalities.}}
\label{table::ablation study input modal}
\setlength{\tabcolsep}{3.5pt}
\resizebox{1\linewidth}{!}{
\begin{tabular}{ccccccccc}
\toprule
\multicolumn{1}{c}{Row} & \multicolumn{2}{c}{Modalities}   & \multicolumn{3}{c}{S2T} & \multicolumn{3}{c}{T2S} \\ 
\cmidrule(lr){2-3} \cmidrule(lr){4-6} \cmidrule(lr){7-9} 
              & Image & Point &  RR@1   &  RR@5  & NDCG@5       & RR@1   &  RR@5  & NDCG@5      \\ \midrule
1 & \ding{51} &    \ding{56}   &    22.97&48.94&16.92&12.99&32.58&22.87
      \\
2 &  \ding{56} & \ding{51}      &   15.68 & 38.14 & 11.70 & 9.22 & 25.24 & 17.34     \\
\rowcolor{orange!15} \cellcolor{white}3 & \ding{51} & \ding{51}      &    \textbf{27.25}    &   \textbf{52.91}   &    \textbf{19.64}      & \textbf{16.18}   & \textbf{35.96}   &   \textbf{26.19}      \\
\bottomrule
\end{tabular}
}
\vspace{-3.5em}
\end{table}

\begin{table}[t]     
\scriptsize

\centering
\caption{\textbf{Ablation study on loss terms.}}
\label{table::ablation study loss terms}
\setlength{\tabcolsep}{3.5pt}
\resizebox{1\linewidth}{!}{
\begin{tabular}{cccccccccc}
\toprule
\multicolumn{1}{c}{Row} & \multicolumn{1}{c}{Retrieval} & \multicolumn{2}{c}{Reconstruction}   & \multicolumn{3}{c}{S2T} & \multicolumn{3}{c}{T2S} \\ 
\cmidrule(lr){3-4} \cmidrule(lr){5-7} \cmidrule(lr){8-10} 
          &  $\mathcal L_{S\leftrightarrow T}$  & $\mathcal L_{rec}^{I \rightarrow P}$ & $\mathcal L_{rec}^{P \rightarrow I}$ &  RR@1   &  RR@5  & NDCG@5       & RR@1   &  RR@5  & NDCG@5      \\ \midrule
1 & \ding{51} &    \ding{56} & \ding{56} &    24.44 &  50.83 &  18.03 &  14.56 &  34.83 &  24.62 \\
2 &  \ding{51} & \ding{51}   & \ding{56}  &   25.76 & 51.54 & 18.67 & 14.96 & 35.07 & 25.13     \\
3 &  \ding{51} & \ding{56}   & \ding{51}  &   25.80 & 51.78 & 18.71 & 14.84 & 35.15 & 25.05     \\
\rowcolor{orange!15} \cellcolor{white}4 & \ding{51} & \ding{51} & \ding{51}      &    \textbf{27.25}    &   \textbf{52.91}   &    \textbf{19.64}      & \textbf{16.18}   & \textbf{35.96}   &   \textbf{26.19}      \\
\bottomrule
\end{tabular}
}
\vspace{-1.5em}
\end{table}

\begin{figure}[t]
\vspace{-0.75em}
\centering
%\captionsetup{justification=centering} % 设置居中对齐
\setlength{\abovecaptionskip}{0pt} % 减少标题上方的间距
\setlength{\intextsep}{0pt} % 减少文本与图片之间的间距
\setlength{\textfloatsep}{0pt} % 减少文本与浮动对象之间的间距

\resizebox{1\linewidth}{!}{
  \subfloat{\hspace{-2.8mm} \includegraphics[width=0.50\linewidth]{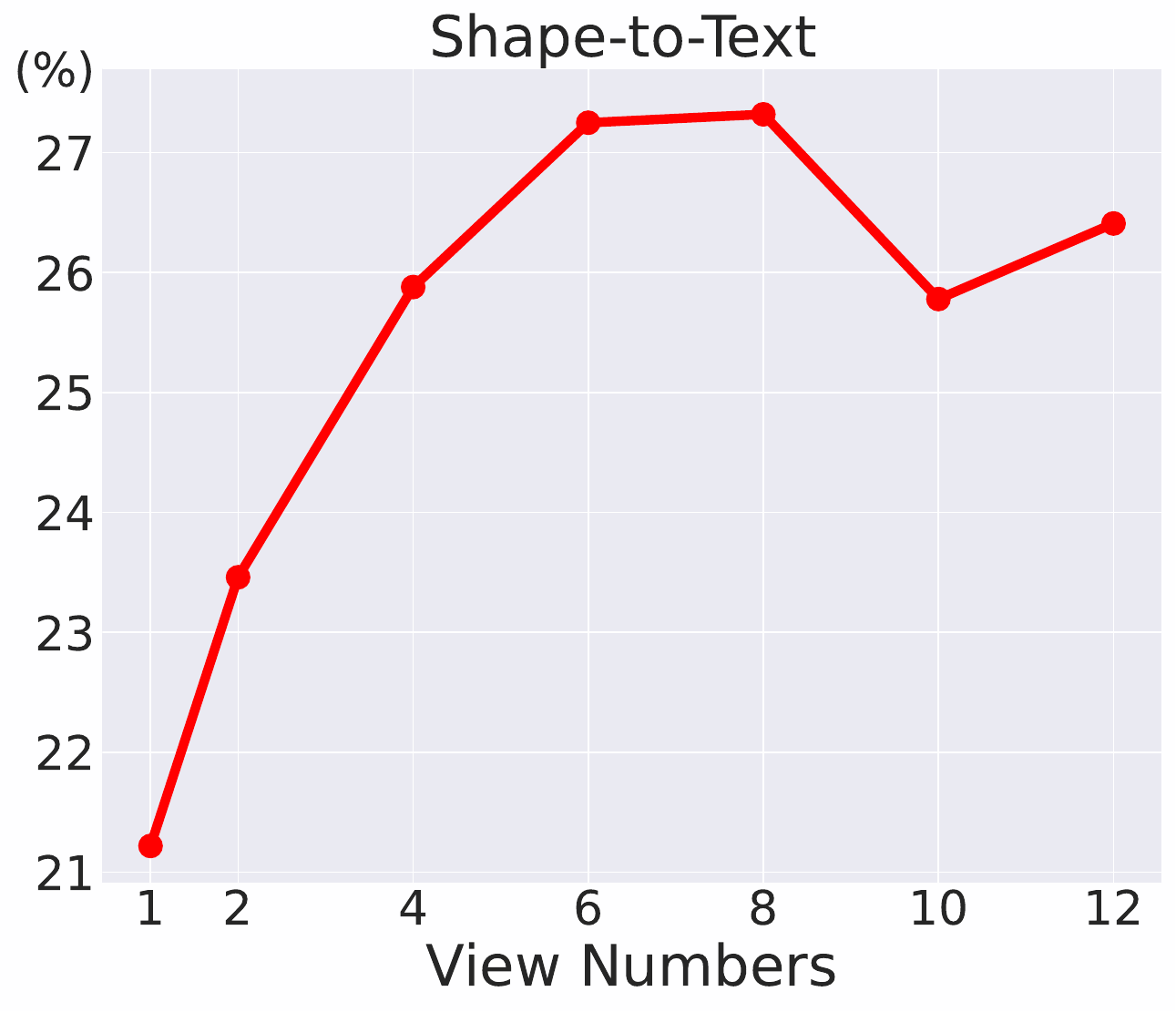}}
  \hfill
  \subfloat{\includegraphics[width=0.50\linewidth]{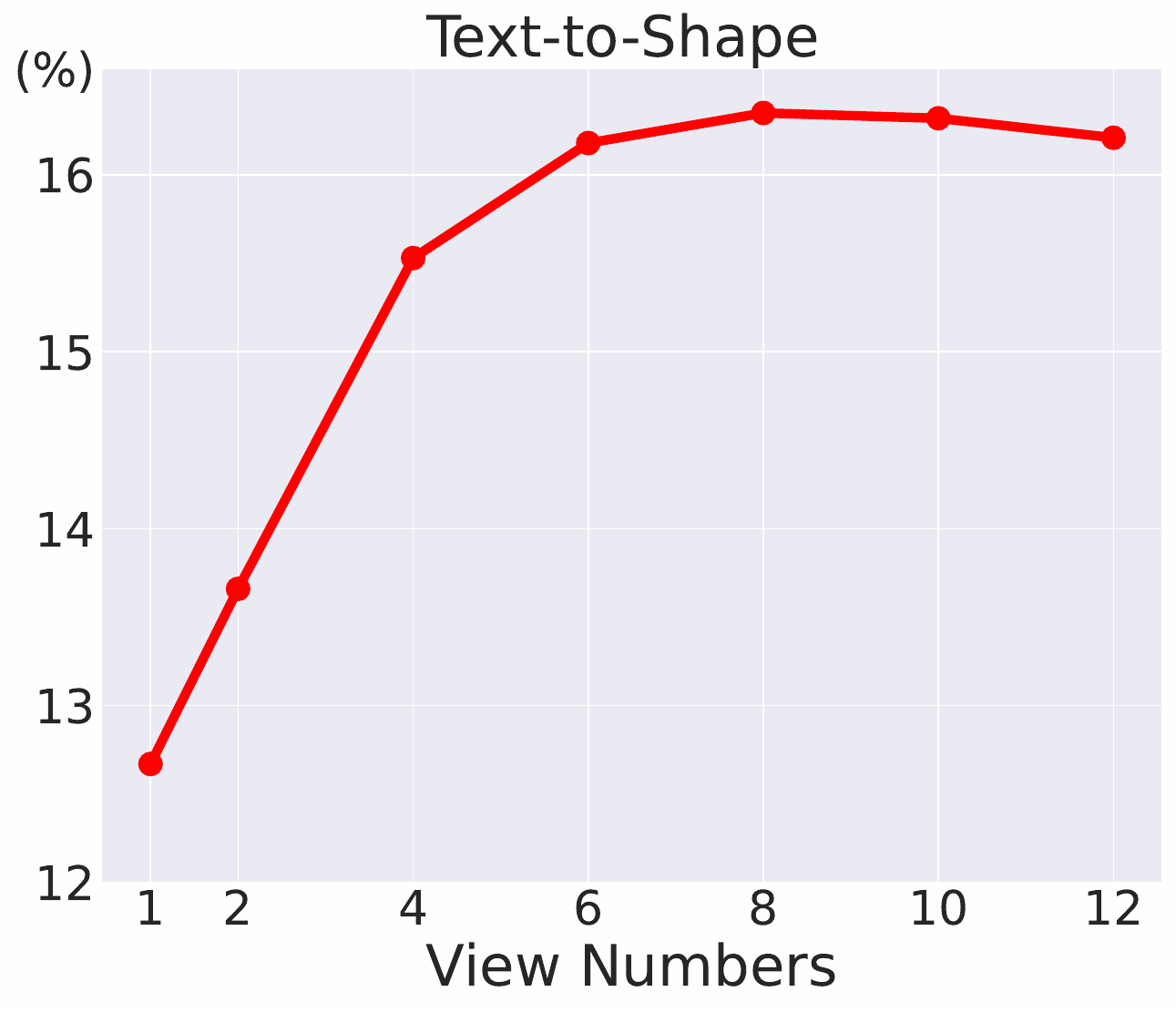}}
}
\caption{\textbf{Ablation study on view numbers with RR@1.}}
\label{fig::ablation study views}
\vspace{-1.5em}
\end{figure}

\paragraph{Evaluation Metrics}
We evaluate the cross-modal 3D retrieval task with the commonly adopted Recall Rate at $k$ (RR@$k$) and Normalized Discounted Cumulative Gain (NDCG). 
RR@k measures the proportion of relevant items successfully retrieved within the top-$k$ results, with $k$ set to \{1, 5\}. 
NDCG assesses the ranking quality by considering both the relevance and position of retrieved items.

\paragraph{Implementation Details}
We employ the ViT-L/14 checkpoint as the pre-trained CLIP model. For point clouds, $n_{p}$ is set to 8,192 and the dimension $d_{p}$ is 6 (xyzrgb). The number of point features $N$ is set to 512.
Each 3D shape is rendered to $M=6$ multi-view images at distinct camera positions, following \cite{ruan2024tricolo} for a fair comparison.
The feature dimension $D$ is set to 1024. Each adapter is an MLP with the ReLU activation function.
For HN-NCE, $\beta$ is set to 0.5 and $\tau$ is initialized to 0.07 following \cite{Radenovic_2023_CVPR}.
The model is trained for 40 epochs with a batch size of 1024. AdamW optimizer \cite{loshchilov2018decoupled} is applied with an initial learning rate of 5e-5 and a cosine annealing schedule.
%\cite{loshchilov2018decoupled}

\subsection{Comparison with State-of-the-Arts}
We compare our method with the following previous state-of-the-art (SOTA) methods: Text2Shape \cite{chen2019text2shape}, $\rm Y^{2}$Seq2Seq \cite{han2019y2seq2seq}, TriCoLo \cite{ruan2024tricolo}, Parts2Words \cite{tang2021parts2words}, and COM3D \cite{wu2024com3d}. The results are borrowed from \cite{wu2024com3d}. The comparison results of shape-to-text and text-to-shape retrieval tasks are presented in Table \ref{table::compare_with_SOTA}. Our method significantly surpasses these methods in all metrics, achieving SOTA results.
Notably, our method remarkably outperforms the previous SOTA method COM3D \cite{wu2024com3d} across all evaluation metrics by a substantial margin, demonstrating the superior effectiveness of our approach.

\subsection{Ablation Studies}

\paragraph{Backbones} For a fair comparison with \cite{ruan2024tricolo, tang2021parts2words, wu2024com3d}, we adopt the same backbone settings of these works in Table \ref{table::ablation study backbone}. We employ trainable MVCNN \cite{su2015multi}, PointNet \cite{qi2017pointnet}, and GRU \cite{chung2014empirical} to encode images, points, and texts, respectively. With this setup, our method still surpasses all prior methods by a large margin, showcasing its remarkable effectiveness.

\paragraph{Input Modalities}
We analyze the impact of input modalities on cross-modal 3D retrieval in Table \ref{table::ablation study input modal}. The retrieval performance declines without point clouds (Row 1). Compared to images, point clouds further capture depth and geometry information, thus contributing to the retrieval task.
Note that the performance significantly drops without image inputs (Row 2). This is due to that it is hard to align point clouds and texts using only MLP adapters. Moreover, the joint application of multi-view images and point clouds leads to substantial improvements. This validates the effectiveness of our reconstruction and multimodal fusion techniques.

\paragraph{Loss Terms}
We assess the impact of loss terms in Table \ref{table::ablation study loss terms}. Without reconstruction losses (Row 1), the performance drops as the generalization ability of encoders for the different modalities declines.
When $\mathcal L_{rec}^{I \rightarrow P}$ or $\mathcal L_{rec}^{P \rightarrow I}$ is employed as a training loss (Row 2 and 3), the performance improves across all metrics. Similarly, the combination of these losses shows further performance improvements, indicating their complementing nature.

\paragraph{Number of Multi-View Images}
We study the effect of the number of views in Fig. \ref{fig::ablation study views}. Increasing the view numbers yields almost consistent enhancements in performance across all metrics. This is because more views lead to a more holistic representation of 3D shapes. In particular, the improvements of adding view numbers from 6 to 12 are limited since 6 views may be enough to represent a holistic 3D shape.

\paragraph{Fusion Scheme}
We verify the effectiveness of the fusion scheme in Table \ref{table::ablation study fusion} Rows 1 and 2. 
The alternative is a simple method that concatenates max-pooled image and point cloud features and then processes with an MLP. The CQA performs better than the MLP fusion as it enables fine-grained cross-modal matching, which better fuses and aligns the embeddings of image and point cloud modalities with essential local geometries and semantics.

\begin{table}[t]     
\scriptsize
\centering
\caption{\textbf{Ablation study on fusion scheme, hard negative contrastive training and reconstruction module.} \textit{MLP} is a simple fusion module with multilayer perceptron and \textit{CQA} represents context-query attention. \textit{Bi-Reconstruct} is the bi-reconstruction method in \cite{feng2023hypergraph} while \textit{Tri-Modal} is our proposed tri-modal reconstruction.
The best results are in \textbf{bold}.}
\label{table::ablation study fusion}
\setlength{\tabcolsep}{3.5pt}
\resizebox{1\linewidth}{!}{
\begin{tabular}{cccccccc}
\toprule
\multicolumn{1}{c}{Row} & \multicolumn{1}{c}{Settings}   & \multicolumn{3}{c}{S2T} & \multicolumn{3}{c}{T2S} \\ 
 \cmidrule(lr){3-5} \cmidrule(lr){6-8} 
      &       &   RR@1   &  RR@5  & NDCG@5       & RR@1   &  RR@5  & NDCG@5      \\ \midrule
\multicolumn{2}{l}{\textit{Fusion Scheme:}}         \\
1& MLP &       24.35 & 50.59 & 17.33 & 14.11 & 34.05 & 24.43     \\
\rowcolor{orange!15} \cellcolor{white}2& CQA      &    \textbf{27.25}    &   \textbf{52.91}   &    \textbf{19.64}      & \textbf{16.18}   & \textbf{35.96}   &   \textbf{26.19}      \\

\midrule
\multicolumn{4}{l}{\textit{Hard Negative Contrastive Training:}}         \\

3& InfoNCE &       23.24 & 50.79 & 17.27 & 14.16 & 34.39 & 24.71 \\
\rowcolor{orange!15} \cellcolor{white}4& HN-NCE      &    \textbf{27.25}    &   \textbf{52.91}   &    \textbf{19.64}      & \textbf{16.18}   & \textbf{35.96}   &   \textbf{26.19}      \\

\midrule

\multicolumn{4}{l}{\textit{Reconstruction Module:}}         \\

5& Bi-Reconstruct \cite{feng2023hypergraph}&       25.42 & 51.76 & 18.34 & 14.70 & 34.99 & 25.17 \\
\rowcolor{orange!15} \cellcolor{white}6& Tri-Modal (Ours)     &    \textbf{27.25}    &   \textbf{52.91}   &    \textbf{19.64}      & \textbf{16.18}   & \textbf{35.96}   &   \textbf{26.19}      \\

\bottomrule
\end{tabular}
}
\vspace{-1em}
\end{table}

\paragraph{Hard Negative Contrastive Training}
We validate the efficacy of hard negative contrastive training (HN-NCE) by replacing it with vanilla InfoNCE. As summarized in Table \ref{table::ablation study fusion} Rows 3 and 4, the retrieval accuracy of vanilla InfoNCE is worse than HN-NCE. With HN-NCE, the model focuses more on the hard negatives during training, leading to more robust cross-modal alignment.

\paragraph{Reconstruction Module}
We compare the impact of our proposed tri-modal reconstruction and the conventional bi-reconstruction
\cite{feng2023hypergraph}. As shown in Table \ref{table::ablation study fusion} Rows 5 and 6, our method leads to better accuracy than bi-reconstruction. Our tri-modal reconstruction facilitates alignment among points, images, and texts, resulting in enhanced generalization ability.

\begin{figure}[t]
\centering
  \subfloat{\includegraphics[width=1\linewidth]{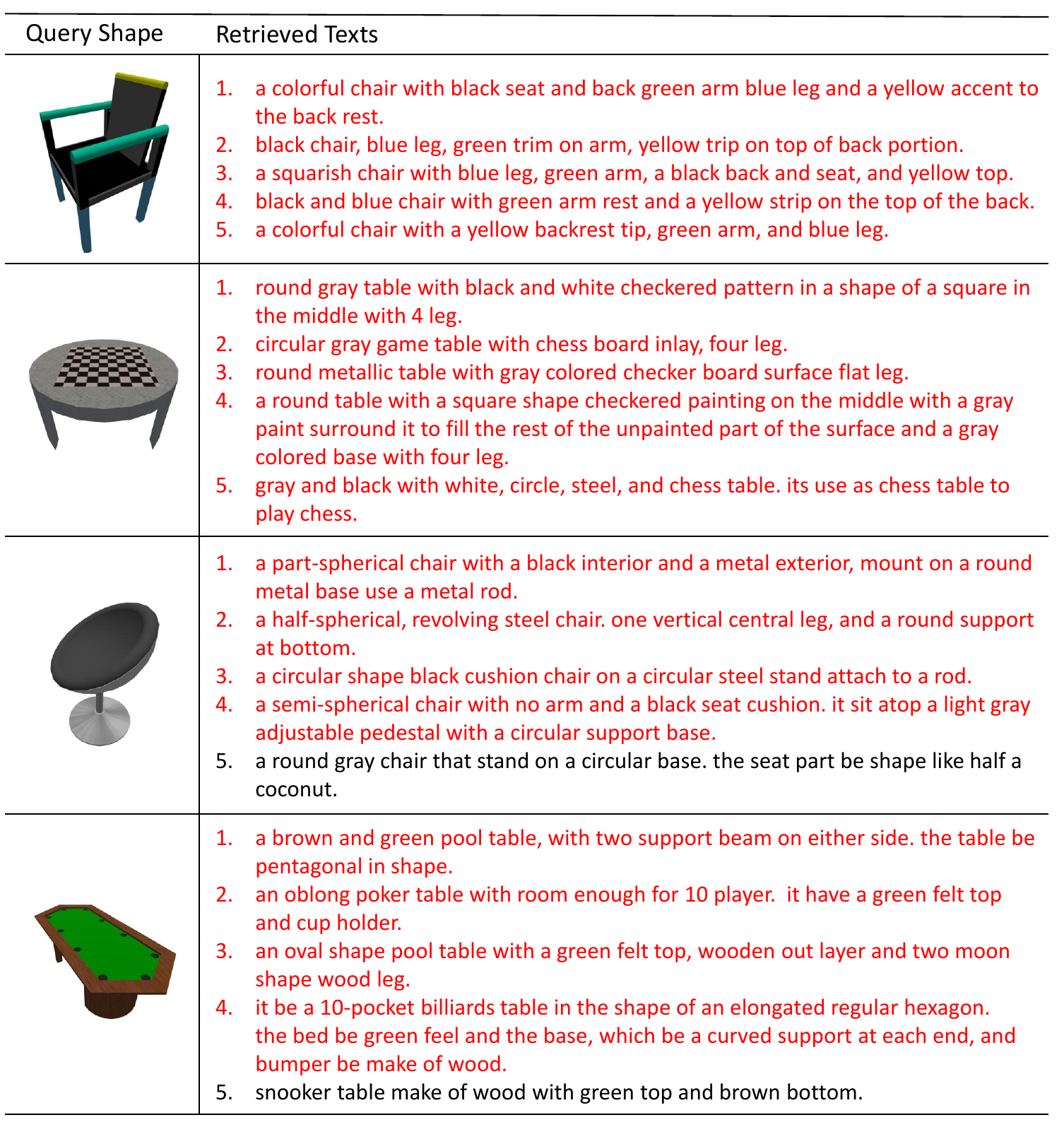}} 
   
\caption{\textbf{Shape-to-text retrieval results.} Each query shape is displayed with the top-5-ranked texts. Ground truths are highlighted in \textcolor[rgb]{1,0,0}{red}. }
\label{figure::Retrieval Results S2T}
\vspace{-1.5em}
\end{figure}

\subsection{Qualitative Results}
\begin{figure}[t]
\centering
  \subfloat{\includegraphics[width=1\linewidth]{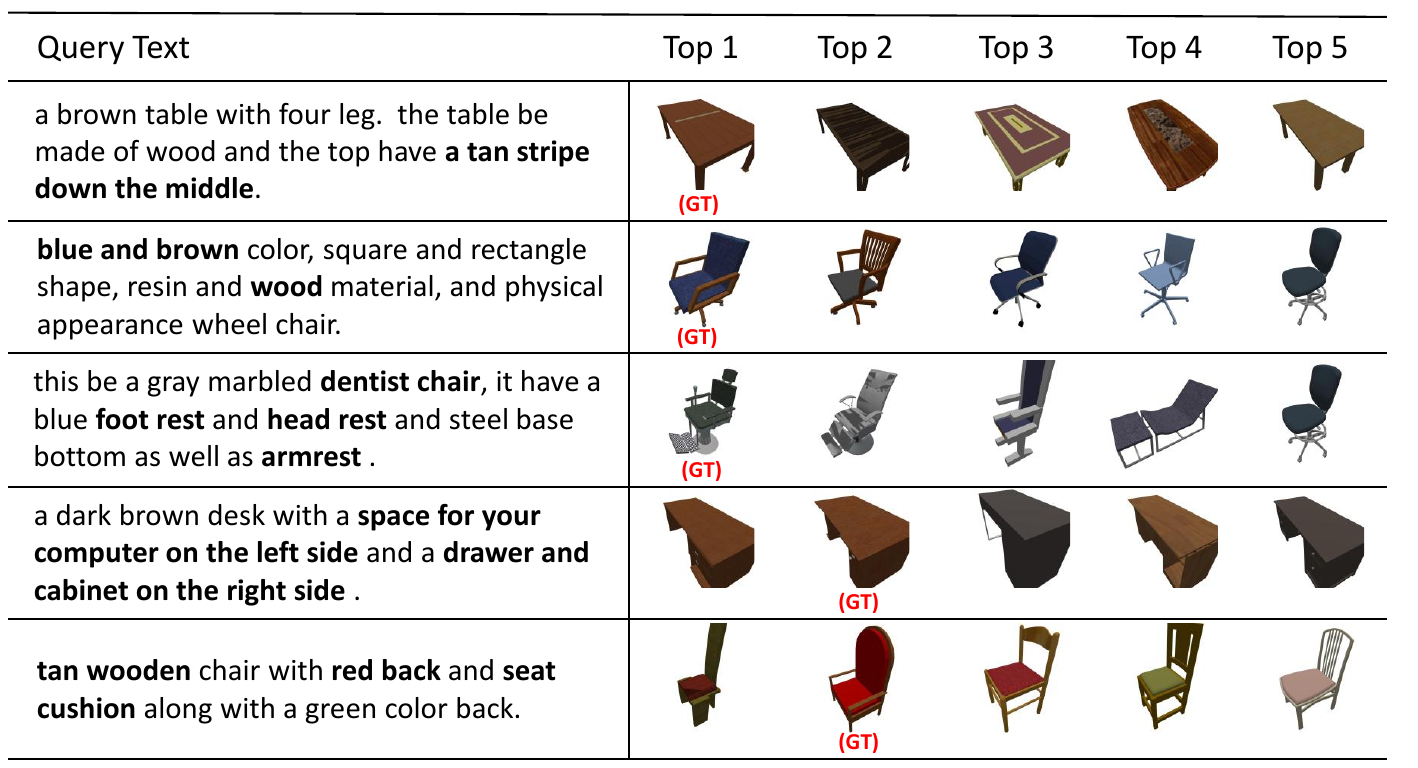}} 
   
\caption{\textbf{Text-to-shape retrieval results.} Each query text is displayed with the top-5-ranked shapes. Words that provide essential details are in \textbf{bold}. Ground truths are as \textcolor[rgb]{1,0,0}{GT}.}
\label{figure::Retrieval Results T2S}
\vspace{-0.5em}
\end{figure}

To qualitatively validate the effectiveness of our method, we report illustrative examples of shape-to-text and text-to-shape retrieval results in Fig. \ref{figure::Retrieval Results S2T} and Fig. \ref{figure::Retrieval Results T2S}, respectively. 
As demonstrated in Fig. \ref{figure::Retrieval Results S2T}, our method accurately retrieves the query shapes with nearly all ground-truth texts (each shape has at least five ground truths). 
Similarly, in Fig. \ref{figure::Retrieval Results T2S}, all retrieved shapes are highly ranked as top 1 or 2, showcasing the remarkable retrieval ability of our method. 
Notably, almost all the retrieved items exhibit comparable semantic characteristics, and even the non-ground-truth results correspond well with the queries. 
We accurately retrieve target items in such a challenging setting where the model must clearly distinguish target items from similar ones, validating the robustness and effectiveness of our method.

\begin{figure}[t]
\centering
  \subfloat{\includegraphics[width=1\linewidth]{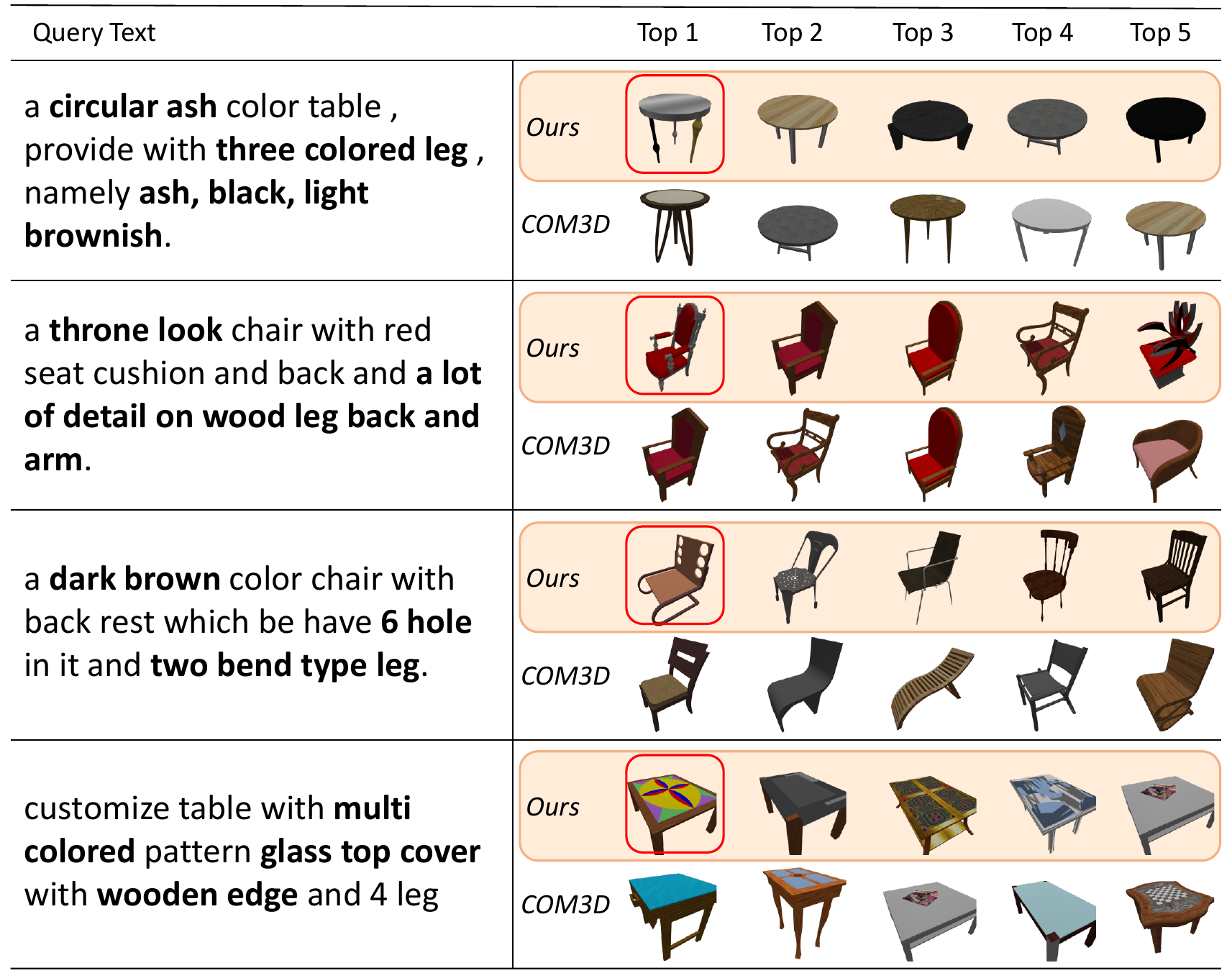}} 
   
\caption{\textbf{Text-to-shape retrieval results of our method and COM3D \cite{wu2024com3d}.} Each query text is displayed with the top-5 ranked shapes. Words that provide essential details are highlighted in \textbf{bold}. Ground truths are marked with \textcolor[rgb]{1,0,0}{red boxes}.}
\label{figure::Retrieval Results T2S Comparision}
\vspace{-1.0em}
\end{figure}

We further compare our method with the previous SOTA method COM3D \cite{wu2024com3d} on the text-to-shape retrieval task. As illustrated in Fig. \ref{figure::Retrieval Results T2S Comparision}, COM3D superficially matches 3D shapes with certain words in the text query. In the first case, the shapes retrieved by COM3D correspond to semantics including ``three colored legs" and specific colors, but none of them meet the condition that the three legs have distinct colors. In contrast, our method accurately retrieves the target shape with correct semantics. The remaining examples also show a similar phenomenon, demonstrating the robust and superior retrieval capability of our method.

\section{Conclusion}

In this paper, we introduce a novel cross-modal 3D retrieval model that bridges the gap between 2D-3D-text tri-modal data. By adopting the tri-modal reconstruction module, we facilitate alignment among point, image, and text modalities. We also leverage 2D-3D consistency and complementary relationships through fine-grained 2D-3D fusion to achieve effective geometric and semantic learning.
Last but not least, we implement hard negative contrastive training to eliminate the noise within datasets, thereby learning discriminative embeddings. 
Extensive experiments on the Text2Shape dataset substantiate the effectiveness of our method in shape-to-text and text-to-shape retrieval tasks, achieving state-of-the-art results.

\vspace{0.5em}

\bibliographystyle{ieeetr}
\bibliography{icme2025references}

\begin{thebibliography}{10}

\bibitem{chen2019text2shape}
K.~Chen, C.~B. Choy, M.~Savva, A.~X. Chang, T.~Funkhouser, and S.~Savarese, ``Text2shape: Generating shapes from natural language by learning joint embeddings,'' in {\em ACCV}, 2018.

\bibitem{han2019y2seq2seq}
Z.~Han, M.~Shang, X.~Wang, {\em et~al.}, ``Y2seq2seq: Cross-modal representation learning for 3d shape and text by joint reconstruction and prediction of view and word sequences,'' in {\em AAAI}, 2019.

\bibitem{ruan2024tricolo}
Y.~Ruan, H.-H. Lee, Y.~Zhang, K.~Zhang, and A.~X. Chang, ``Tricolo: Trimodal contrastive loss for text to shape retrieval,'' in {\em WACV}, 2024.

\bibitem{tang2021parts2words}
C.~Tang, X.~Yang, B.~Wu, Z.~Han, and Y.~Chang, ``Parts2words: Learning joint embedding of point clouds and texts by bidirectional matching between parts and words,'' in {\em CVPR}, 2023.

\bibitem{wu2024com3d}
H.~Wu, R.~Li, H.~Wang, {\em et~al.}, ``Com3d: Leveraging cross-view correspondence and cross-modal mining for 3d retrieval,'' in {\em ICME}, 2024.

\bibitem{schuhmann2022laion5b}
C.~Schuhmann, R.~Beaumont, R.~Vencu, C.~Gordon, R.~Wightman, {\em et~al.}, ``Laion-5b: An open large-scale dataset for training next generation image-text models,'' {\em NeurIPS}, 2022.

\bibitem{kakaobrain2022coyo-700m}
M.~Byeon, B.~Park, H.~Kim, S.~Lee, {\em et~al.}, ``Coyo-700m: Image-text pair dataset.'' \url{https://github.com/kakaobrain/coyo-dataset}, 2022.

\bibitem{miech2019howto100m}
A.~Miech, D.~Zhukov, J.-B. Alayrac, M.~Tapaswi, I.~Laptev, and J.~Sivic, ``Howto100m: Learning a text-video embedding by watching hundred million narrated video clips,'' in {\em ICCV}, 2019.

\bibitem{radford2021learning}
A.~Radford, J.~W. Kim, C.~Hallacy, {\em et~al.}, ``Learning transferable visual models from natural language supervision,'' in {\em ICML}, 2021.

\bibitem{li2023blip2}
J.~Li, D.~Li, {\em et~al.}, ``Blip-2: Bootstrapping language-image pre-training with frozen image encoders and large language models,'' in {\em ICML}, 2023.

\bibitem{huang2023language}
S.~Huang, L.~Dong, W.~Wang, Y.~Hao, S.~Singhal, S.~Ma, {\em et~al.}, ``Language is not all you need: Aligning perception with language models,'' {\em NeurIPS}, 2023.

\bibitem{liu2024visual}
H.~Liu, C.~Li, {\em et~al.}, ``Visual instruction tuning,'' {\em NeurIPS}, 2023.

\bibitem{li-etal-2024-groundinggpt}
Z.~Li, Q.~Xu, D.~Zhang, H.~Song, Y.~Cai, {\em et~al.}, ``{G}rounding{GPT}: Language enhanced multi-modal grounding model,'' in {\em ACL}, 2024.

\bibitem{fang2023eva}
Y.~Fang, W.~Wang, B.~Xie, Q.~Sun, L.~Wu, {\em et~al.}, ``Eva: Exploring the limits of masked visual representation learning at scale,'' in {\em CVPR}, 2023.

\bibitem{li2022blip}
J.~Li, D.~Li, {\em et~al.}, ``Blip: Bootstrapping language-image pre-training for unified vision-language understanding and generation,'' in {\em ICML}, 2022.

\bibitem{zhai2023sigmoid}
X.~Zhai, B.~Mustafa, A.~Kolesnikov, and L.~Beyer, ``Sigmoid loss for language image pre-training,'' in {\em ICCV}, 2023.

\bibitem{chen2023altclip}
Z.~Chen, G.~Liu, {\em et~al.}, ``Altclip: Altering the language encoder in clip for extended language capabilities,'' in {\em Findings of ACL}, 2023.

\bibitem{chang2015shapenet}
A.~X. Chang, T.~Funkhouser, L.~Guibas, P.~Hanrahan, {\em et~al.}, ``Shapenet: An information-rich 3d model repository,'' {\em arXiv:1512.03012}, 2015.

\bibitem{tran2015learning}
D.~Tran, L.~Bourdev, R.~Fergus, L.~Torresani, and M.~Paluri, ``Learning spatiotemporal features with 3d convolutional networks,'' in {\em ICCV}, 2015.

\bibitem{chung2014empirical}
J.~Chung, C.~Gulcehre, K.~Cho, and Y.~Bengio, ``Empirical evaluation of gated recurrent neural networks on sequence modeling,'' in {\em NIPS 2014 Workshop on Deep Learning, December 2014}, 2014.

\bibitem{sajjadi2022scene}
M.~S. Sajjadi, H.~Meyer, E.~Pot, U.~Bergmann, K.~Greff, {\em et~al.}, ``Scene representation transformer: Geometry-free novel view synthesis through set-latent scene representations,'' in {\em CVPR}, 2022.

\bibitem{yu2022point}
X.~Yu, L.~Tang, Y.~Rao, T.~Huang, {\em et~al.}, ``Point-bert: Pre-training 3d point cloud transformers with masked point modeling,'' in {\em CVPR}, 2022.

\bibitem{feng2023hypergraph}
Y.~Feng, S.~Ji, Y.-S. Liu, S.~Du, {\em et~al.}, ``Hypergraph-based multi-modal representation for open-set 3d object retrieval,'' {\em TPAMI}, 2023.

\bibitem{yu2018qanet}
A.~W. Yu, D.~Dohan, M.-T. Luong, R.~Zhao, K.~Chen, {\em et~al.}, ``Qanet: Combining local convolution with global self-attention for reading comprehension,'' in {\em ICLR}, 2018.

\bibitem{seo2016bidirectional}
M.~Seo, A.~Kembhavi, A.~Farhadi, and H.~Hajishirzi, ``Bidirectional attention flow for machine comprehension,'' in {\em ICLR}, 2017.

\bibitem{Radenovic_2023_CVPR}
F.~Radenovic, A.~Dubey, A.~Kadian, T.~Mihaylov, S.~Vandenhende, {\em et~al.}, ``Filtering, distillation, and hard negatives for vision-language pre-training,'' in {\em CVPR}, 2023.

\bibitem{oord2018representation}
A.~v.~d. Oord, Y.~Li, and O.~Vinyals, ``Representation learning with contrastive predictive coding,'' {\em arXiv preprint arXiv:1807.03748}, 2018.

\bibitem{loshchilov2018decoupled}
I.~Loshchilov and F.~Hutter, ``Decoupled weight decay regularization,'' in {\em ICLR}, 2019.

\bibitem{su2015multi}
H.~Su, S.~Maji, E.~Kalogerakis, {\em et~al.}, ``Multi-view convolutional neural networks for 3d shape recognition,'' in {\em ICCV}, 2015.

\bibitem{qi2017pointnet}
C.~R. Qi, H.~Su, K.~Mo, and L.~J. Guibas, ``Pointnet: Deep learning on point sets for 3d classification and segmentation,'' in {\em CVPR}, 2017.

\end{thebibliography}

\end{document}